\pdfoutput=1 
\documentclass[10pt,twocolumn,letterpaper]{article}

\usepackage{cvpr}              

\usepackage{graphicx}
\usepackage{amsmath}
\usepackage{amssymb}
\usepackage{booktabs}
\usepackage{array}
\usepackage{multirow}
\usepackage{stfloats}
\usepackage{stfloats}
\usepackage[accsupp]{axessibility}

\usepackage[pagebackref,breaklinks,colorlinks]{hyperref}

\usepackage[capitalize]{cleveref}
\crefname{section}{Sec.}{Secs.}
\Crefname{section}{Section}{Sections}
\Crefname{table}{Table}{Tables}
\crefname{table}{Tab.}{Tabs.}

\def\cvprPaperID{10407} 
\def\confName{CVPR}
\def\confYear{2022}

\usepackage{enumitem}
\setenumerate[1]{itemsep=0pt,partopsep=0pt,parsep=\parskip,topsep=5pt}
\setitemize[1]{itemsep=0pt,partopsep=0pt,parsep=\parskip,topsep=5pt}
\setdescription{itemsep=0pt,partopsep=0pt,parsep=\parskip,topsep=5pt}

\usepackage{color}
\usepackage{flushend}
\usepackage{setspace}
 
\begin{document}

\title{DAIR-V2X: A Large-Scale Dataset for Vehicle-Infrastructure Cooperative \\ 3D Object Detection}
\author{Haibao Yu$^{1}$, Yizhen Luo$^{1,3}$, Mao Shu$^{2}$, Yiyi Huo$^{1,4}$, Zebang Yang$^{1,3}$, Yifeng Shi$^{2}$, Zhenglong Guo$^{2}$,
\and Hanyu Li$^{2}$, Xing Hu$^{2}$, Jirui Yuan$^{1}$, Zaiqing Nie$^{1}$\thanks{Corresponding author. $^{3,4}$ Work done while at AIR.} \\
$^{1}$Institute for AI Industry Research(AIR), Tsinghua University \\
$^{2}$ Baidu Inc. $^{3}$ Department of Computer Science and Technology, Tsinghua University\\$^{4}$ University of Chinese Academy of Science \\
{\tt\small \{yuhaibao@air.,luoyz18@mails.,yzb19@mails.,yuanjirui@air.,zaiqing@air.\}tsinghua.edu.cn,} \\ 
{\tt\small \{shumao,shiyifeng,guozhenglong,lihanyu02,huxing\}@baidu.com,  huoyiyi18@mails.ucas.ac.cn} 
}

\maketitle

\begin{abstract}
Autonomous driving faces great safety challenges for a lack of global perspective and the limitation of long-range perception capabilities. 
It has been widely agreed that vehicle-infrastructure cooperation is required to achieve Level 5 autonomy. 
However, there is still NO dataset from real scenarios available for computer vision researchers to work on vehicle-infrastructure cooperation-related problems. To accelerate computer vision research and innovation for Vehicle-Infrastructure Cooperative Autonomous Driving (VICAD), we release DAIR-V2X Dataset, which is the first large-scale, multi-modality, multi-view dataset from real scenarios for VICAD. 
DAIR-V2X comprises 71254 LiDAR frames and 71254 Camera frames, and all frames are captured from real scenes with 3D annotations.
The Vehicle-Infrastructure Cooperative 3D Object Detection problem (VIC3D) is introduced, formulating the problem of collaboratively locating and identifying 3D objects using sensory inputs from both vehicle and infrastructure. 
In addition to solving traditional 3D object detection problems, the solution of VIC3D needs to consider the temporal asynchrony problem between vehicle and infrastructure sensors and the data transmission cost between them. 
Furthermore, we propose Time Compensation Late Fusion (TCLF), a late fusion framework for the VIC3D task as a benchmark based on DAIR-V2X.
Find data, code, and more up-to-date information
at \href{https://thudair.baai.ac.cn/index}{https://thudair.baai.ac.cn/index} and \href{https://github.com/AIR-THU/DAIR-V2X}{https://github.com/AIR-THU/DAIR-V2X}.
\end{abstract}

\section{Introduction}
Autonomous driving (AD) is arguably one of the hottest topics currently occupying public attention and imagination. 
The success of deep neural networks brings the promise of solving AD's core requirement to perceive the surrounding environment from point cloud~\cite{lang2019pointpillars, Yang2020ssd,shi2019pointrcnn}, images~\cite{rukhovich2021imvoxelnet,reading2021categorical} or multi-modality data~\cite{Vora2020Point,Sindagi2019MVX}.
\begin{figure}[htbp]
	\centering
	\includegraphics[width=0.5\textwidth]{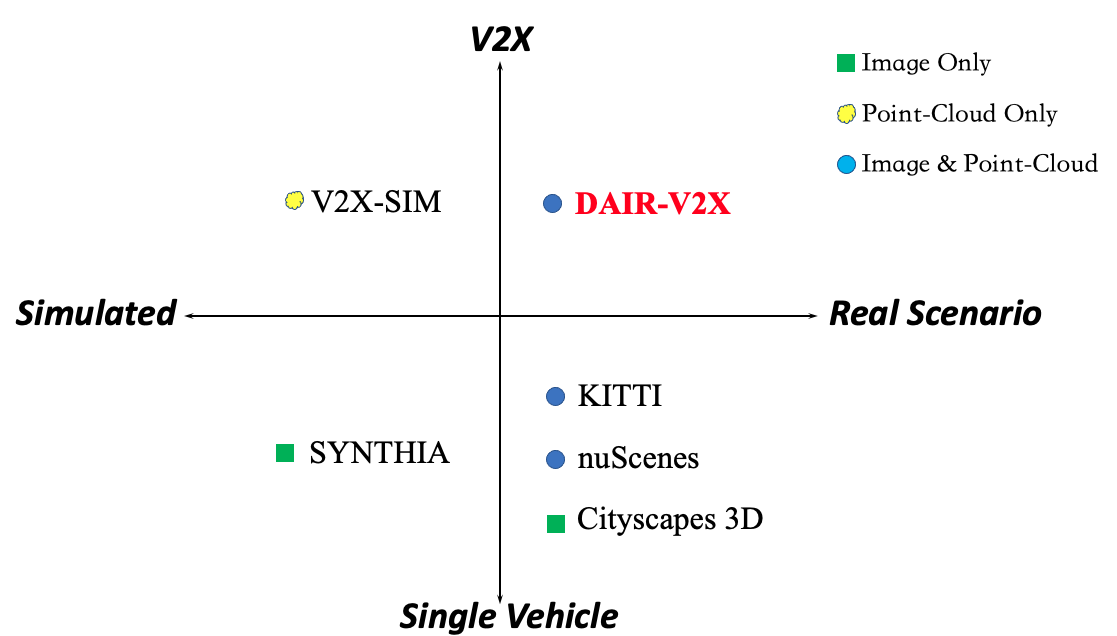}
	\caption{Datasets available for 3D Object Detection in autonomous driving.
DAIR-V2X is the first real-world V2X dataset for VICAD.}
	\label{fig:positive_effect_vic3d}
\end{figure}
Despite its great progress recently, autonomous driving still faces great safety challenges for a lack of global perspective and the limitation of long-range perception capability. It has been widely agreed that vehicle-infrastructure cooperation is required to achieve Level 5 autonomy. Utilizing both vehicle and infrastructure sensors brings a number of significant advantages, including providing a global perspective far beyond the current horizon and covering blind spots.
Advances in communications like V2X (vehicle to everything) have made it possible to utilize data from infrastructure sensors~\cite{chen2017vehicle,storck20195g}.
However, there is still NO dataset from real scenarios available for researchers to work on vehicle-infrastructure cooperation-related problems.

To accelerate computer vision research and innovation for Vehicle-Infrastructure Cooperative Autonomous Driving (VICAD), we release DAIR-V2X Dataset, which is the first large-scale, multi-modality, multi-view dataset for VICAD.
It contains 71254 LiDAR frames and 71254 Camera frames captured in intersection scenes where a well-equipped vehicle passes through intersections with infrastructure sensors deployed. 40\% of the frames are captured from infrastructure sensors and 60\% of the frames are captured from vehicle sensors. All of them are precisely labeled by expert annotators. 
The dataset covers 10 km of city roads, 10 km of highway, 28 intersections, and 38 km$^2$ of driving regions with diverse weather and lighting variations. More details could be found in Tab.~\ref{tab: dataset comparison}.

In this paper, the Vehicle-Infrastructure Cooperative 3D Object Detection (VIC3D) task is introduced, formulating the problem of cooperatively locating and identifying 3D objects using sensory inputs from both vehicle and infrastructure. In addition to solving traditional 3D object detection problems, the solution of VIC3D needs to consider the temporal asynchrony problem and data transmission cost between vehicle and infrastructure sensors. 

\begin{table*}[ht]
\caption{A detailed comparison between autonomous driving-related datasets.  - indicates that specific information is not provided. In particular, DAIR-V2X is composed of DAIR-V2X-C, DAIR-V2X-V and DAIR-V2X-I, where DAIR-V2X-C is captured by both vehicle and infrastructure sensors, DAIR-V2X-V is captured by vehicle sensors, and DAIR-V2X-I is captured by infrastructure sensors.}
\label{tab: dataset comparison}
\resizebox{\textwidth}{!}{%
\begin{tabular}{lccccccc}
\hline
\hline
\textbf{Dataset} & \textbf{Year} & \textbf{Real/Simulated} & \textbf{View} & \textbf{Image} & \textbf{Pointcloud} & \textbf{3D boxes} & \textbf{Classes} \\
\hline
KITTI\cite{Geiger2012KITTI}  & 2012 & real   & single vehicle                       & 15k & 15k & 200k & 8  \\
\hline
nuScenes\cite{Caesar2020NuScenes}      & 2019 & real      & single vehicle                       & 1.4M & 400k & 1.4M & 23 \\
\hline
Waymo Open\cite{sun2020scalability}  & 2019 & real      & single vehicle                       & 1M & 200k & 12M  & 4  \\
\hline
ApolloScape\cite{huang2019apolloscape}   & 2018 & real & single vehicle & 144k & 0 & 70k & 8-35 \\
\hline
BBD100K\cite{yu2020bdd100k}       & 2020 & real      & single vehicle                       & 100M & 0 & 0  & 10 \\
\hline
ONCE\cite{mao2021one} & 2021 & real      & single vehicle                       & 7M & 1M & 417k  & 5 \\
\hline
SYNTHIA\cite{ros2016synthia}       & 2016 & simulated  & single vehicle                       & 213k & 0 & - & 13  \\
\hline
V2X-Sim\cite{li2021learning}       & 2021 & simulated & multi-vehicle                        & 0 & 10k & 26.6k & 2  \\
\hline
highD\cite{krajewski2018highd}         & 2018 & real      & infrastructure (UAV)                                  & 1.53M & 0 & 0   & 1  \\
\hline
\hline
\textbf{DAIR-V2X (Our)}      & 2021 & real      & vehicle-infrastructure cooperative & 71k & 71k & 1.2M & 10   \\
\hline
 - \textbf{ DAIR-V2X-C}    & 2021 & real      & vehicle-infrastructure cooperative & 39k & 39k & 464k & 10 \\
\hline
 - \textbf{ DAIR-V2X-V}    & 2021 & real      & single vehicle                       & 22k & 22k & 239k & 10  \\
\hline
 - \textbf{ DAIR-V2X-I}    & 2021 & real      & infrastructure               & 10k & 10k & 493k & 10   \\
\hline
\hline
\end{tabular}%
}
\end{table*}

To resolve the VIC3D object detection task and facilitate future research, we also introduce our VIC3D object detection benchmark in this paper.
For data with less temporal asynchrony problems, we implement both early fusion and late fusion approaches. Results show that the average precision of fusion methods is 10 to 20 points higher than detectors that only use information from a single view. 
Results also show that early fusion can achieve better performance than late fusion but requires more data transmission. With the DAIR-V2X dataset, we expect more future research to achieve a performance-bandwidth trade-off.
For data with severe temporal asynchrony, we propose a Time Compensation Late Fusion framework, which can effectively alleviate the temporal asynchrony problem.

The key contributions of our work are as follows:
\begin{itemize}
    \item We release the DAIR-V2X dataset, which is the first large-scale dataset for vehicle-infrastructure cooperative autonomous driving. All frames are captured from real scenarios with 3D annotations.
    \item We formulate the problem of cooperatively locating and identifying 3D objects using sensory inputs from both vehicle and infrastructure as VIC3D.
    \item We introduce benchmarks for VIC3D object detection and single-view 3D object detection tasks.
    The results show the effectiveness of vehicle-infrastructure cooperation in VIC3D object detection.
    Especially, we propose the Time Compensation Late Fusion framework to alleviate the temporal asynchrony problem. 
\end{itemize}

\section{Relative Work}
\begin{figure*}[ht!]
	\centering
	\includegraphics[width=1\textwidth]{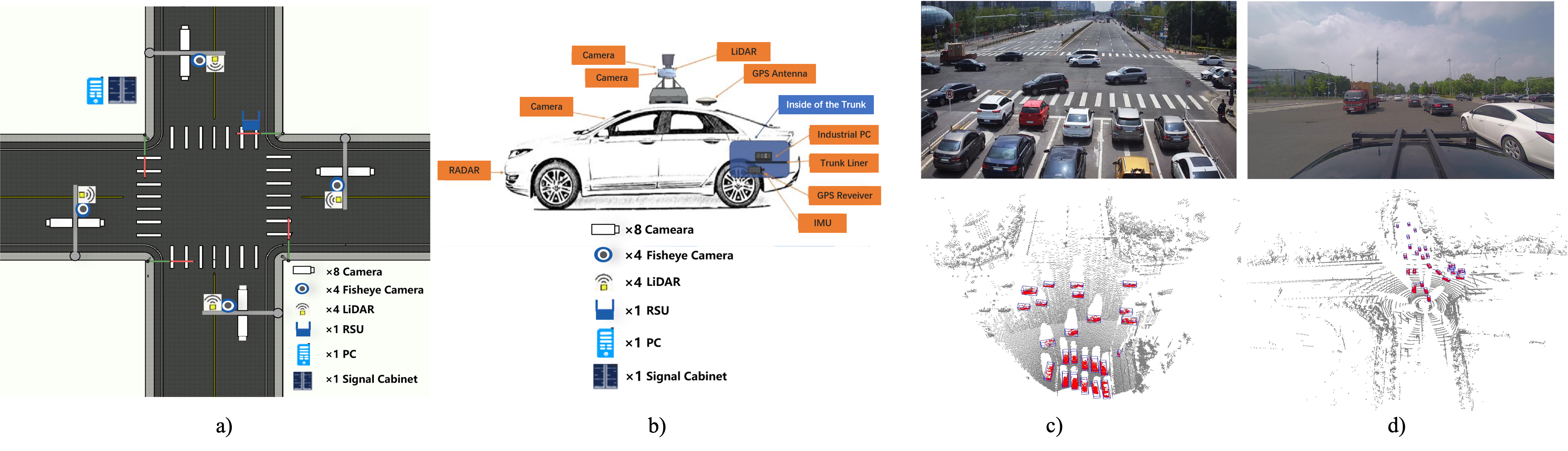}
	\caption{a) Acquisition system with infrastructure sensors. b) Acquisition system with vehicle sensors. c) Infrastructure-view image and point cloud with 3d annotation. Paired vehicle-view and infrastructure-view information complement each other in the perspective of view. d) Vehicle-view image and point cloud with 3d annotation.}
	\label{fig:equipment}
\end{figure*}
\subsection{Autonomous Driving Datasets}
In recent years, an increasing number of autonomous driving datasets have been released and greatly promoted the development of autonomous driving research. Datasets like SYNTHIA ~\cite{ros2016synthia} and Cityscapes ~\cite{cordts2016cityscapes} mainly focus on 2D annotations for images. KITTI~\cite{Geiger2012KITTI} and nuScenes~\cite{Caesar2020NuScenes} are multi-modality datasets providing camera images as well as LiDAR point clouds. Nevertheless, all datasets mentioned above only provide data from a single-vehicle view. V2X-SIM~\cite{li2021learning} is an attempt to generate a multi-vehicle view dataset, but the dataset was generated by a simulator rather than captured from real scenarios. Compared with those datasets, our DAIR-V2X dataset is the first large-scale, multi-modality, multi-view dataset captured from real scenarios for VICAD, and contains data captured from the Vehicle-Infrastructure Cooperative view.
Tab.~\ref{tab: dataset comparison} shows the comparison of our dataset with the others. 
In our DAIR-V2X, we also provide a Repo3D~\cite{ye2022rope3d} dataset composed of multi-source infrastructure images and 3D annotations, for those who are interested in Mono3D object detection and domain adaptation.

\subsection{3D Detection}
3D object detection serves as the prerequisite for the success of autonomous driving. Many techniques have been introduced and can be roughly classified into three categories. a) Image-based 3D Detection refers to methods that detect 3D objects directly from 2D images. ImVoxelNet\cite{rukhovich2021imvoxelnet} is a good example to make predictions from images. 
b) Pointcloud-based 3D Detection stands for manners that make 3D object detection merely from point clouds. PointPillars~\cite{lang2019pointpillars}, SECOND~\cite{yan2018second}, and 3DSSD~\cite{Yang2020ssd} are such approaches that achieve convincing detection results from point clouds. 
c) Multimodality-based 3D Detection uses both images and point clouds to make predictions. Pointpainting~\cite{Vora2020Point} and MVXNet~\cite{Sindagi2019MVX} are practices of fusing image and LiDAR features to predict 3D bounding boxes.
While 3D object detection has made great progress recently, there are still some tough problems that remain to solve such as blind spots and weak  long-distance perception. 
To explore how to utilize the infrastructure information to solve the problems mentioned above, 
we conduct VIC3D object detection based on our dataset proposed in this paper.

\subsection{Multi-Sensor Fusion}

Multi-sensor fusion\cite{Wang2020Multi} is the integration of heterogeneous information collected by different sensors to alleviate the uncertainty and vulnerability of systems that rely on a single sensor.
Based on the fusion stage, multi-sensor fusion can be categorized into early fusion, intermediate fusion, and late fusion. a) In early fusion, raw data from different sensors are directly transferred and fused~\cite{gao2018early}. b) In intermediate fusion, intermediate representations like features extracted from the models are fused~\cite{Chen2017Multi, Sindagi2019MVX}. c) In late fusion, the prediction outputs like 3D information of the objects are fused~\cite{hag2018late}. 
VIC3D can be considered as a variant of the multi-sensor problem, so previous fusion methods can be taken into consideration to integrate the infrastructure information.
However, in addition to the multi-sensor fusion challenges, VIC3D faces difficulties caused by the temporal asynchrony problem and the data transmission constraint.

\subsection{V2X Cooperative Perception}
V2X aims to build a communication system between vehicles and other devices in a complex traffic environment. Current V2X research mainly focuses on V2V (Vehicle-to-Vehicle) and V2I (Vehicle-to-Infrastructure) area. V2VNet\cite{Wang2020V2VNetVC} is a pioneering work in V2V that broadcasts compressed intermediate features and propagates message received from nearby vehicles to generate motion forecasts.  Works of V2I~\cite{Cui2019Automatic, ZHAO2019Detection} leverage infrastructure LiDAR data to generate and broadcast detection results. 
However, none of these approaches have been verified on a dataset captured from real scenarios. This may cause a huge gap between theory and practice. Therefore, we release the DAIR-V2X dataset to boost further study in this field.

\section{The DAIR-V2X Dataset}
In order to facilitate research on VICAD, we release DAIR-V2X, a large-scale, multi-modality, multi-view dataset from real scenarios with 3D annotations for vehicle infrastructure cooperation.
Here we describe how we set up infrastructure and vehicle sensors, select interesting scenes, annotate the dataset and protect the privacy of third parties.


\subsection{Setup}
\noindent \textbf{Equipment.}
Equipment for data collection are composed of infrastructure sensors and vehicle sensors.
a) Infrastructure sensors. Each of the 28 intersections selected from Beijing High-level Autonomous Driving Demonstration Area are deployed with four pairs of 300-beam LiDAR and high-resolution camera. 
The DAIR-V2X dataset picks only one pair of them.
b) Vehicle sensors. One 40-beam LiDAR and one high-quality camera looking forward are mounted on top of the autonomous vehicles. 
Specific layout is posted in Figure \ref{fig:equipment}, and precise details are displayed in Table \ref{tab: sensors}.
\begin{table}
\caption{Key Sensor Specifications in DAIR-V2X. Veh. stands for vehicle view, and Inf. stands for infrastructure view.}
\label{tab: sensors}
\small
\begin{tabular}{lp{5.4cm}}
\hline
\textbf{Sensor} & \textbf{Details} \\
\hline
Inf. LiDAR & 300 beams, 10Hz capture frequency, $100^o$ horizontal FOV, $-30^o$ to $10^o$ vertical FOV, $\leq 280m$ range, $\pm 3cm$ accuracy  \\
Inf. Camera & RGB, 25Hz capture frequency, 1920x1080 resolution, JPEG compressed \\
\hline
Veh. LiDAR & 40 beams, 10Hz capture frequency, $360^{o}$ horizontal FOV, $-30^o$ to $10^o$ vertical FOV, $\leq 200m$ range, $\pm 0.33^{o}$ vertical resolution \\
Veh. Camera & RGB, 20Hz capture frequency, 1920x1080 resolution, JPEG compressed \\
Veh. GPS \& IMU & 1000HZ update rate \\
\hline
\end{tabular}
\end{table} \\

\noindent \textbf{Coordinate.}
There are 5 types of coordinate systems on DAIR-V2X, $i.e.$, the LiDAR coordinate, the camera coordinate, the image coordinate, the world coordinate, and the positioning coordinate. 
The origin of the LiDAR coordinate system is located at the center of the LiDAR sensor, the x-axis is positive forwards, the y-axis is positive to the left, and the z-axis is positive upwards.
The infrastructure LiDAR coordinate system is converted from its original system which has an inclination angle with the ground. 
The real-time relative pose of the equipped vehicle is obtained from GPS/IMU combined with SLAM and a local map. There is also manual secondary labeling confirmation to ensure calibration accuracy.
The Lidar-to-Camera transformation is obtained by multiplying Lidar-to-World and World-to-Camera transformations.


\subsection{Data Acquisition}\label{sec:DataAcquisition}
\noindent \textbf{Collection.}
We drive a well-equipped vehicle in the collection area and save the corresponding vehicle frames and infrastructure frames respectively.
After the collection of raw data, we manually select 100 representative scenes of 20s duration.
Such scenes include vehicle data and infrastructure data, where vehicles drive through intersections deployed with equipment.
We sample key frames at 10Hz from both sides to form DAIR-V2X-C.
In DAIR-V2X-C, it is important to note that the timestamp difference between a vehicle frame and its closest infrastructure frame could be slightly varied, due to the asynchronous triggering between vehicle sensors and infrastructure sensors. 
We sample 22K frames from additionally about 350 vehicle-only segments of 60s duration to form DAIR-V2X-V, and sample 10K frames from additionally about 150 infrastructure-only segments duration to form DAIR-V2X-I, to enlarge the dataset.
Compared to the single-view data in DAIR-V2X-
C, DAIR-V2X-V and DAIR-V2X-I contains more diverse scenes and will be more challenging to only improve the single-view performance.

\noindent \textbf{Annotation.}
With multiple validation steps and refinement processes, expert annotators make high-quality annotations for infrastructure frames and vehicle frames respectively.
Specifically, annotators exhaustively label each of the 10 object classes in every image and point cloud frame with its category attribute, occlusion state, truncated state, and a 7-dimensional cuboid modeled as $x, y, z, width, length, height$, and $yaw\ angle$. 
10 categories include different vehicles, pedestrian, different cyclists. 
Moreover, experts also meticulously annotate objects in camera images with a rectangle bounding box modeled as $x, y, width$, and $length$. \\
\indent To be mentioned, we also conduct semi-automatic labeling for the cooperative annotations with vehicle and infrastructure frame pairs.
We first select vehicle and infrastructure frame pairs from DAIR-V2X-C. The timestamp differences between the two frames of the selected pairs are less than $10ms$ (We call it the Synchronous Case which is defined in Section~\ref{sec: vic3d object detection}. To obtain more cooperative annotations, we 
extend the threshold from $10ms$ to $30ms$).
Next, we convert infrastructure 3D boxes into vehicle LiDAR coordinate system and fuse the vehicle annotations and infrastructure annotations. 
For each 3D box in the infrastructure annotation, if we can not find any 3D box in the vehicle annotation that has the same location and category, we add the infrastructure 3D box into the vehicle annotations; in this way, we get the vehicle-infrastructure cooperative annotations.
We manually supervise and adjust the cooperative annotations to generate more accurate annotations.
Here we take 9331 infrastructure frames and vehicle frames as well as the cooperative annotations to form the VIC-Sync dataset for our VIC3D object detection benchmark. 

\noindent \textbf{Protection.}
The whole dataset is desensitized before public release. Complied with local laws and regulations, we erase all localization information, including road name, map data, and positioning information, to make sure our dataset meets requirements. In addition, we utilize professional labeling tools to blur all the information suspected of privacy violation, including road signs, license plates and faces, to protect  privacy and avoid violating personal rights.

\section{Task \& Metrics}
Autonomous driving faces great safety challenges for a lack of global perspective and the limitation of long-range perception capabilities. 
Since 3D object detection is one of the key perception tasks in autonomous driving, in this paper, we focus on the vehicle-infrastructure cooperative (VIC) 3D object detection task, the vehicle receives and integrates information from infrastructure to localize and recognize objects surrounding itself.
Compared with traditional multi-sensor 3D object detection tasks, VIC3D object detection has the following alternative characteristics:
\begin{itemize}
\item Transmission Cost. Limited by physical communication conditions, fewer data should be transmitted from infrastructure to reduce bandwidth consumption, alleviate time delay, and satisfy real-time requirements. Thus, the solution to VIC3D object detection needs to balance the trade-off between the performance and the transmission cost.
\item Temporal Asynchrony. Timestamps of data from the vehicle sensors and the infrastructure sensors are different due to the asynchronous triggering and time delay caused by transmission cost, to generate the temporal-spatial error. Therefore, temporal synchronization should be considered in solving VIC3D.
\end{itemize}

To better formulate the VIC3D object detection task, we will give a detailed definition to the VIC3D object detection and then provide two metrics to measure detection performance and transmission cost in this section. 

\subsection{VIC3D Object Detection} \label{sec: vic3d object detection}
VIC3D object detection can be formulated as the optimization problem of effectively integrating infrastructure and vehicle information to localize and recognize 3D objects considering transmission cost.
Here we discuss what the input and output of VIC3D should be. \\

\noindent \textbf{Input.}
The input of VIC3D is composed of data from the vehicle and the infrastructure. 
\begin{itemize}
    \item Vehicle Frame $I_{v}(t_{v})$: captured at time $t_{v}$ as well as its relative pose $M_{v}(t_{v})$, where $I_{v}(\cdot)$ denotes the capturing function of vehicle sensors.
    \item Infrastructure Frame $I_{i}(t_{i})$: captured at time $t_{i}$ as well as its relative pose $M_{v}(t_{i})$, where $I_{i}(\cdot)$ denotes the capturing function of infrastructure sensors.
\end{itemize}
Note that $t_{i}$ should be earlier than $t_{v}$ because there is a time delay caused by data transmission from the infrastructure to the vehicle.
Considering that the objects would move so slightly in the tiny time interval that the spatial offset can be ignored, we take the case that $|t_v-t_i|\le 10ms$ as Synchronous Case (i.e. $t_v\approx t_i$). 
Similarly, we take the case that $|t_v-t_i|>10ms$ as Asynchronous Case.
In addition, we allow using more infrastructure frames previous to $I_{i}(t_{i})$ in solving VIC3D to make full use of the infrastructure computing resources. \\

\noindent \textbf{Ground Truth.}
The outputs of VIC3D object detection contain 3D information like the location, category, and orientation of objects surrounding the vehicle. 
The corresponding ground truth of VIC3D is the fusion result of infrastructure and vehicle ground truth, which could be formulated as:
\begin{equation}
    GT = GT_{v} \cup GT_{i},
\end{equation}
where $GT_{v}$ is the ground truth for vehicle sensor perception and $GT_{i}$ is the ground truth for infrastructure sensor perception.

VIC3D is mainly used to improve the perception performance of the self-driving vehicle.
We are more concerned about a certain range of egocentric surroundings and the 3D information of objects at time $t_v$ than at $t_i$.
Therefore, $GT_{v}$ and $GT_{i}$ should both be based on time $t_v$.
However, the timestamp of the input frames captured from the infrastructure and captured from the vehicle could be different that $t_v \neq t_i$.
This not only brings challenges to fusing the infrastructure information in model prediction but also creates huge problems to generate the ground truth.
That's because objects annotated with infrastructure frame at time $t_i$ may move to different locations at time $t_v$, and we cannot directly get the infrastructure frame at time $t_v$ to annotate.

In response to these difficulties, we discuss how we generate the ground truth for VIC3D based on DAIR-V2X.
\begin{itemize}
    \item Synchronous Case (i.e. $t_{v}\approx t_{i}$).
    Under this condition, an object that appears in vehicle frame $I_{v}(t_{v})$ should have the same spatial location as it appears in infrastructure frame $I_{i}(t_{i})$.
    Therefore, we can directly take the vehicle-infrastructure cooperative 3D annotations obtained by semi-automatic labeling illustrated in Section~\ref{sec:DataAcquisition} as ground truth.
    \item Asynchronous Case (i.e. $t_{v} \neq t_{i}$).
    If we can find such infrastructure frame $I_i(t_i')$ satisfying $|t_v-t_{i}^{'}|\le 10ms$, we can generate ground truth with $I_i(t_{i}^{'})$.
    If not, we have to estimate the 3D states of objects at $t_v$ to generate ground truth.
    This work can be carried out based on the tracking ID and kinematic equation after we provide the tracking ID in future work.
\end{itemize}

\subsection{Evaluation Metrics.}
VIC3D object detection has two major goals: better detection performance and less transmission cost.
We describe the metrics for such two goals below. \\

\noindent\textbf{Average Precision.}
$AP$ (Average precision) is a popular metric for measuring the object detectors performance~\cite{everingham2010pascal}.
We also use $AP$ to evaluate the 3d detection performance with cooperative annotations as ground truth.
Since we are more concerned about egocentric surroundings, we remove objects outside the designed area.
Here we set the designed area as a rectangular area as [0, -39.12, 100, 39.12].

\noindent\textbf{Transmission Cost.}
We use $AB$ (Average Byte) to measure the transmission cost.
Here Byte is a unit of digital information that consists of eight $bit$s. 
To simplify the problem, we ignore the time consumption of data encoders and decoders during transmission.
That means the less transmission cost, the less time delay.
Data to be transmitted from the infrastructure can be one or a combination of the following forms.
\begin{itemize}
    \item Raw data such as images or point clouds contains complete information but requires much transmission cost. 
    \item Intermediate representation requires less transmission cost while retaining valuable information, which may achieve a better performance-transmission trade-off. 
    Surely, this requires a more sophisticated design to extract suitable intermediate representation.
    \item Object-level outputs directly provide 3D object information. Although it is transmission-efficient, it may lose valuable information.
    \item Other auxiliary information like scene flows help to alleviate temporal asynchrony problems. 
\end{itemize}

\section{Benchmark}
In this section, we provide a VIC3D object detection benchmark and a Single-View (SV) 3D object detection benchmark on our DAIR-V2X dataset, analyze their characteristics and suggest avenues for future research.

\begin{table*}[htpb!]
\centering
\small
\caption{VIC3D object detection Benchmark on DAIR-V2X-C.}
\label{tab: VIC3D object detection results}
 \scalebox{0.80}{
\begin{tabular}{cccc|ccccccccc}
\hline
\hline
\multirow{2}{*}{Modality} & 
\multirow{2}{*}{Fusion} & 
\multirow{2}{*}{Model} & 
\multirow{2}{*}{Dataset} & 
\multicolumn{4}{c}{$AP_{3D(IoU=0.5)}$} & 
\multicolumn{4}{c}{$AP_{BEV(IoU=0.5)}$} & 
$AB$ \\ \cline{5-8} \cline{9-12}
& &   &  & 
Overall & 0-30m & 30-50m & 50-100m & 
Overall & 0-30m & 30-50m & 50-100m &  ($Byte$) \\
\hline\hline
\multirow{3}{*}{Image} & 
  Veh.-Only & ImvoxelNet~\cite{rukhovich2021imvoxelnet} & VIC-Sync &
  12.03 & 16.25 & 7.25 & 2.28 & 13.62 & 17.66 & 8.58 & 2.82 & 0 \\
  & Inf.-Only & ImvoxelNet~\cite{rukhovich2021imvoxelnet} & VIC-Sync &
  19.93 & 27.34 & 17.61 & 14.43 & 25.31 & 32.02 & 23.28 & 20.38 & 102.32 \\
  & Late Fusion & ImvoxelNet~\cite{rukhovich2021imvoxelnet} & VIC-Sync &
  26.56 & 34.20 & 17.20 & 9.81 & 31.40 & 37.75 & 21.21 & 12.99 & 102.32 \\
\hline\hline

\multirow{4}{*}{Pointcloud} & 
  Veh.-Only & PointPillars~\cite{lang2019pointpillars} & VIC-Sync &
  31.33 & 27.48 & 25.58 & 12.63 & 35.06 & 30.55 & 28.65 & 14.16 & 0 \\
  & Inf.-Only & PointPillars~\cite{lang2019pointpillars} & VIC-Sync &
  17.62 & 16.54 & 10.98 & 9.17 & 24.40 & 21.47 & 16.00 & 13.07 & 336.16 \\
  & Late Fusion & PointPillars~\cite{lang2019pointpillars} & VIC-Sync &
  41.90 & 37.65 & 32.72 & 18.84 & 47.96 & 42.40 & 37.65 & 22.08 & 336.16 \\
  & Early Fusion & PointPillars~\cite{lang2019pointpillars} & VIC-Sync & 50.03 & 53.07 & 60.38 & 33.05 & 53.73 & 55.80 & 64.08 & 36.17  & 1382275.75 \\
\hline\hline

\multirow{3}{*}{Pointcloud} & 
  Late Fusion & PointPillars~\cite{lang2019pointpillars} & VIC-Async-$1$ &
  40.21 & 34.17 & 29.40 & 15.50 & 46.41 & 38.05 & 34.10 & 19.20 & 341.08 \\
  & Late Fusion & PointPillars~\cite{lang2019pointpillars} & VIC-Async-$2$ &
  35.29 & 32.16 & 28.07 & 13.44 & 40.65 & 35.62 & 32.35 & 15.88 & 306.79 \\
  & Early Fusion & PointPillars~\cite{lang2019pointpillars} & VIC-Async-$1$ & 47.47 & 48.88 & 58.86 & 30.89 & 51.67 & 52.70 & 63.09 & 34.72 & 1362216.0 \\
\hline\hline

\multirow{2}{*}{Pointcloud} &
TCLF & PointPillars~\cite{lang2019pointpillars} & VIC-Async-$1$ &
  40.79 & 34.67 & 29.69 & 15.76 & 46.80 & 38.24 & 34.27 & 19.40 & 539.60 \\
  & TCLF & PointPillars~\cite{lang2019pointpillars} & VIC-Async-$2$ &
  36.72 & 33.91 & 29.41 & 14.52 & 41.67 & 36.78 & 33.36 & 17.18 & 506.70 \\
\hline\hline

\hline
\end{tabular}
}
\end{table*}

\subsection{Benchmark for VIC3D object detection}
We provide a benchmark for VIC3D object detection on the VIC-Sync dataset extracted from DAIR-V2X-C, which is illustrated in Section~\ref{sec:DataAcquisition}. The dataset is composed of 9311 pairs of infrastructure and vehicle frames as well as their cooperative annotations as ground truth. 
Besides, we take the temporal asynchrony between the infrastructure frame and the vehicle frame into consideration in the benchmark, which is mainly caused by the difference in the sampling rate and transmission delay.
To simulate the temporal asynchrony phenomenon, we replace each infrastructure frame in the VIC-Sync dataset with the infrastructure frame which is $k$-th frame previous to the original infrastructure frame to construct the VIC-Async-$k$ dataset for the benchmark. In our experiments, we set $k=1, 2$.
We split VIC-Sync and VIC-Async-$k$ datasets to train/valid/test part as 5:2:3 respectively.
We use cooperative annotations to evaluate the detection results under the vehicle-egocentric view.
The experiment results are presented in Table~\ref{tab: VIC3D object detection results}.

\subsubsection{Baselines}\label{sec: vic3d baselines}
Here we present several baselines with different modalities and fusion methods for VIC3D object detection.

\paragraph{LiDAR detection baseline with Late Fusion.}
To demonstrate the performance improvement by utilizing both infrastructure and vehicle data, we implement a late fusion framework with an infrastructure detector and a vehicle detector.
Firstly, we choose PointPillars~\cite{lang2019pointpillars} as the 3D detector and train the two detectors with  infrastructure-view and vehicle-view data in VIC-Sync separately.
Then, we convert the infrastructure predictions into the vehicle LiDAR coordinate system and merge the prediction results with a matcher based on the Euclidian distance measurement and the Hungarian method\cite{Kuhn2010TheHM} to generate fusion results. 

\begin{figure*}[htpb]
    \centering
    \includegraphics[width=0.9\textwidth]{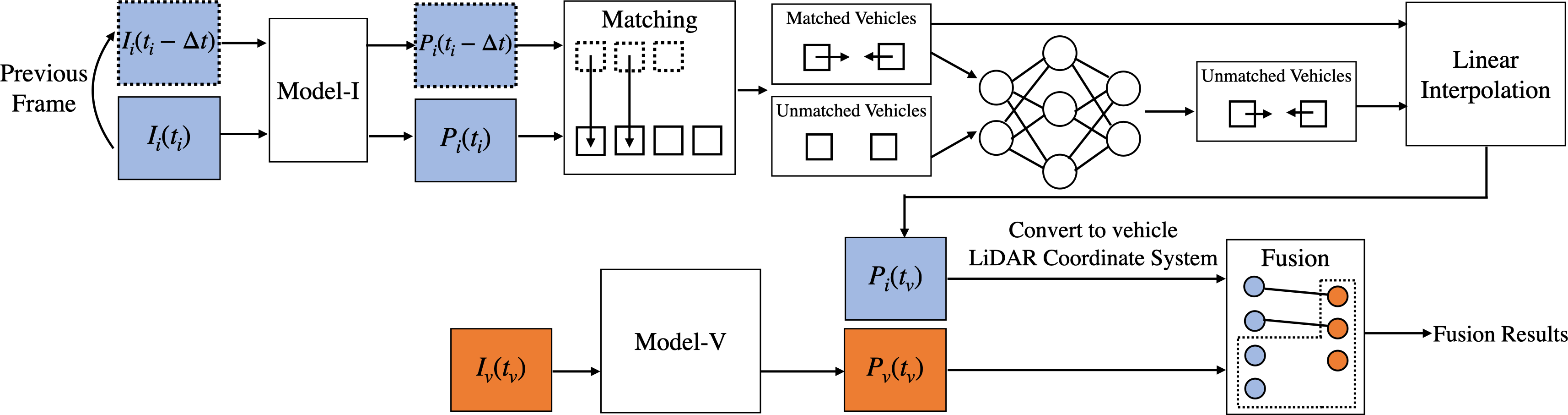}
    \caption{Time Compensation Late Fusion (TCLF) Framework. $\Delta t$ denotes the sampling interval of infrastructure sensors. We predict and match the boxes between two infrastructure frames. For matched vehicles, we compute their velocities directly. For unmatched vehicles, we feed the position and motion information of the current scene into an MLP to predict their velocities. Finally, we can approximate the positions of vehicles at $t_{v}$ by linear interpolation, and fuse the results of the vehicle frame.}
    \label{fig:late fusion framework}
\end{figure*}

To illustrate the temporal asynchrony problem, we also implement the LiDAR detection late fusion baseline on the VIC-Async-$k$ dataset.
In addition, based on tracking and state estimation we propose the Time Compensation Late Fusion (TCLF) framework.
The TCLF is mainly composed of the following three parts: 
1) Estimating the velocity of the objects with two adjacent infrastructure frames. 
2) Estimating the state of the infrastructure objects at $t_{v}$.
3) Fusing the estimated infrastructure predictions and vehicle predictions following the way of LiDAR late fusion baseline.
The details of the TCLF framework could be seen in Fig. \ref{fig:late fusion framework}. 

Note that we also report the evaluation results only with the infrastructure data and only with the vehicle data, which are named as Veh.-Only and Inf.-Only respectively.
The evaluation results are presented in Tab.~\ref{tab: VIC3D object detection results}. 

\paragraph{Image detection baseline with Late Fusion.}
To examine image-only VIC3D object detection, we also implement the late fusion framework only with infrastructure images and vehicle images. 
We choose ImvoxelNet~\cite{rukhovich2021imvoxelnet} as the 3D detector and train infrastructure detector and vehicle detector with the corresponding part of VIC-Sync training data separately.
We implement the image detection late fusion following the LiDAR detection late fusion.

\paragraph{LiDAR detection baseline with Early Fusion.}
To explore the fusion effect at the raw data level, we implement the early fusion with PointPillars~\cite{lang2019pointpillars} as the 3D detector on the VIC-Sync dataset.
We first convert the infrastructure point cloud in the VIC-Sync dataset into the vehicle LiDAR coordinate system, then fuse the infrastructure point cloud and vehicle point cloud.
We directly train and evaluate the detector with the fused point cloud.
Further to illustrate the temporal asynchrony problem, we also implement the early fusion with PointPillars~\cite{lang2019pointpillars} on the VIC-Async-$k$ dataset.

\begin{figure}[htpb]
    \centering
    \includegraphics[width=0.4\textwidth]{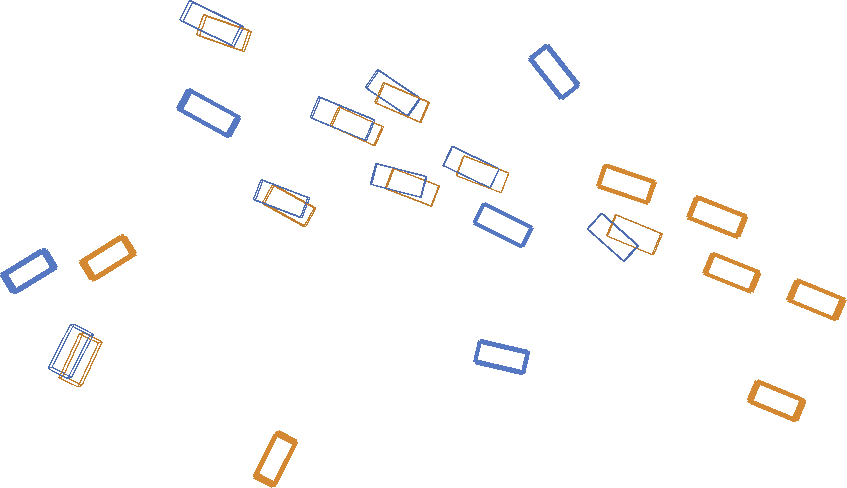}
    \caption{Prediction results of the vehicle frame (orange) and infrastructure frame (blue). We observe that infrastructure data (thick blue boxes) supplements the blind spot and extends the perception field for the vehicle.}
    \label{fig: fusion vis}
\end{figure}

\begin{figure}[htpb]
    \centering
    \includegraphics[width=0.3\textwidth]{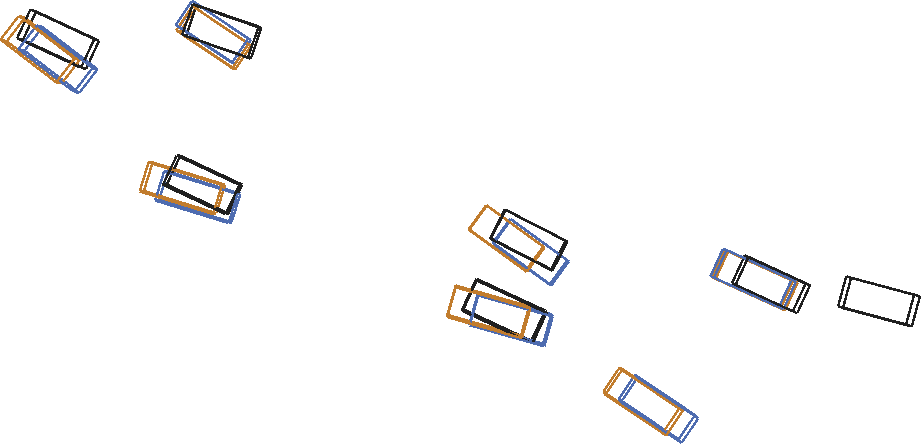}
    \caption{Prediction results with and without time compensation. The results of TCLF (blue) have a larger overlap with ground truth (black) than the results without time compensation (orange).}
    \label{fig: time comp vis}
\end{figure}

\subsubsection{Analysis}
Here we analyze the properties of the methods for the VIC3D object detection benchmark in Section~\ref{sec: vic3d baselines}.

\paragraph{Cooperative-view vs. Single-view.} 
We compare the performance of the methods whether using both infrastructure data and vehicle data. 
In Tab.~\ref{tab: VIC3D object detection results}, the detection performance of late fusion is much better than the performance of Veh.-Only or Inf.-Only, whether it is Image-based or LiDAR-based or it is based on VIC-Sync dataset or VIC-Async-$k$ dataset.
For example, the LiDAR detection with Late Fusion achieves overall 41.90 AP points for 3D detection and overall 47.96 AP points for BEV detection on the VIC-Sync dataset.
However, the LiDAR detection only with vehicle data just achieves overall 31.33\%  AP for 3D detection and overall 35.06\%  AP for BEV detection, and the LiDAR detection only with infrastructure data just achieves overall 17.62\% AP for 3D detection and overall 24.40\% AP for BEV detection.
The experiment results demonstrate that fusing the infrastructure information can effectively improve the perception performance of the vehicle. 
This is mainly because infrastructure data provides supplementary information that makes up for the vehicle's perception field.
A visualization example is shown in Fig. \ref{fig: fusion vis}.

\paragraph{Temporal Asynchrony vs Time Compensation.} 
Temporal asynchrony brings challenges to fusing the infrastructure data. 
Compared with the results on the VIC-Sync dataset, the performance of LiDAR detection with fusion drops significantly on VIC-Async-$k$ (2 points on VIC-Async-$1$ and 6 points on VIC-Async-$2$). 
The decline is mainly due to the state changes of moving objects, resulting in matching difficulties and fusion errors.
However, our TCLF can effectively improve the performance of late fusion up to 0.5\% AP and 1.5\% AP on VIC-Async-$1$ and VIC-Async-$2$ respectively, which demonstrates that time compensation can effectively alleviate the temporal asynchrony problems especially when the time delay is larger.
A visualization example is provided in Fig.~\ref{fig: time comp vis}.

\paragraph{Early Fusion vs. Late Fusion.} 
Compared with late fusion, early fusion achieves up to 8\% AP higher under both BEV and 3D benchmarks, whether it is based on the VIC-Sync dataset or the VIC-Async-$1$ dataset.
However, early fusion should transmit the whole point cloud and suffers an extremely high transmission cost, which is about 4000 times more than late fusion. 
For more practical applications, we encourage future research on achieving better performance while consuming less transmission bandwidth.
We will also release the feature fusion for the benchmark in the future.

\begin{table*}[htpb!]
\centering
\caption{SV3D Detection Benchmark on DAIR-V2X-V}
\label{tab: SS3D-V}
\small
\begin{tabular}{cc|ccccccccc}
\hline\hline
\multirow{2}{*}{Modality} & 
\multirow{2}{*}{Model} & 
\multicolumn{3}{c}{Vehicle$_{3D(IoU=0.5)}$} & 
\multicolumn{3}{c}{Pedestrian$_{3D(IoU=0.25)}$} & 
\multicolumn{3}{c}{Cyclist$_{3D(IoU=0.25)}$} \\ \cline{3-11} &  
& Easy & \multicolumn{1}{c}{Middle} & \multicolumn{1}{c}{Hard} 
& Easy & \multicolumn{1}{c}{Middle} & \multicolumn{1}{c}{Hard} 
& Easy & \multicolumn{1}{c}{Middle} & \multicolumn{1}{c}{Hard} \\
\hline\hline
Image & ImvoxelNet~\cite{rukhovich2021imvoxelnet} & 38.37 & 24.28 & 21.54 & 4.54 & 4.54 & 4.54 & 10.38 & 9.09 & 9.09 \\
PointCloud & PointPillars~\cite{lang2019pointpillars} & 61.76 & 49.02 & 43.45 & 33.40 & 24.68 & 22.39 & 38.24 & 33.80 & 32.35 \\
PointCloud & SECOND~\cite{yan2018second} & 69.44 & 59.63 & 57.63 & 43.45 & 39.06 & 38.78 & 44.21 & 39.49 & 37.74 \\
Image+PointCloud & MVXNet~\cite{Sindagi2019MVX} & 69.86 & 60.74 & 59.31 & 47.73 & 43.37 & 42.49 & 45.68 & 41.84 & 40.55 \\

\hline\hline
\end{tabular}
\end{table*}

\begin{table*}[htpb!]
\centering
\caption{SV3D Detection Benchmark on DAIR-V2X-I}
\label{tab: SS3D-I}
\small
\begin{tabular}{cc|ccccccccc}
\hline\hline
\multirow{2}{*}{Modality} & 
\multirow{2}{*}{Model} & 
\multicolumn{3}{c}{Vehicle$_{3D(IoU=0.5)}$} & 
\multicolumn{3}{c}{Pedestrian$_{3D(IoU=0.25)}$} & 
\multicolumn{3}{c}{Cyclist$_{3D(IoU=0.25)}$} \\ \cline{3-11} &  
& Easy & \multicolumn{1}{c}{Middle} & \multicolumn{1}{c}{Hard} 
& Easy & \multicolumn{1}{c}{Middle} & \multicolumn{1}{c}{Hard} 
& Easy & \multicolumn{1}{c}{Middle} & \multicolumn{1}{c}{Hard} \\
\hline\hline
Image & ImvoxelNet~\cite{rukhovich2021imvoxelnet} & 44.78 & 37.58 & 37.55 & 6.81 & 6.746 & 6.73 & 21.06 & 13.57 & 13.17 \\
PointCloud & PointPillars~\cite{lang2019pointpillars} & 63.07 & 54.00 & 54.01 & 38.53 & 37.20 & 37.28 & 38.46 & 22.60 & 22.49 \\
PointCloud & SECOND~\cite{yan2018second} & 71.47 & 53.99 & 54.00 & 55.16 & 52.49 & 52.52 & 54.68 & 31.05 & 31.19 \\
Image+PointCloud & MVXNet~\cite{Sindagi2019MVX} & 71.04 & 53.71 & 53.76 & 55.83 & 54.45 & 54.40 & 54.05 & 30.79 & 31.06 \\

\hline\hline
\end{tabular}
\end{table*}

\subsection{Benchmark for SV3D Detection}
We present an extensive 3D detection benchmark for those who are interested in Single-View (SV) 3D detection tasks based on DAIR-V2X-V and DAIR-V2X-I datasets. 
Compared with the single-side data in DAIR-V2X-C, the two datasets are more diverse and could be more challenging to implement 3D object detection.
Hence, we encourage researchers who just aim at improving the performance of vehicle 3D object detection or infrastructure 3D object on DAIR-V2X-V and DAIR-V2X-I.

We split DAIR-V2X-V and DAIR-V2X-I datasets to train/valid/test part as 5:2:3 respectively.
We present a number of baselines with methods based on different modalities on the two datasets respectively:
ImvoxelNet~\cite{rukhovich2021imvoxelnet}, PointPillars~\cite{lang2019pointpillars}, SECOND~\cite{yan2018second} and MVXNet~\cite{Sindagi2019MVX}.
We evaluate 3D object detection performance using the PASCAL criteria as KITTI~\cite{Geiger2012KITTI}, that distant objects are filtered out based on their bounding box height in the image plane. Three types of modes are used for evaluation, including Easy, Moderate, and Hard modes.
We implement these baselines with MMDetection3D Framework~\cite{mmdet3d2020}.
Evaluation results are shown in Tab.~\ref{tab: SS3D-V} and Tab.~\ref{tab: SS3D-I}. 


\section{Conclusion}
In this paper, we introduce DAIR-V2X, the first large-scale, multi-modality, multi-view dataset for vehicle-infrastructure cooperative autonomous driving, and all frames are captured from real scenes with 3D annotations. 
We also define VIC3D object detection to formulate the problem of collaboratively locating and identifying 3D objects using sensory input from both vehicle and infrastructure. 
In addition to solving traditional 3D object detection problems, the solution of VIC3D needs to consider the temporal asynchrony problem between vehicle and infrastructure sensors and the data transmission cost between them.   
To facilitate future research, we provide a VIC3D benchmark for detection models with our proposed Time Compensation Late Fusion framework, as well as extensive benchmarks for 3D detection on vehicle-view and infrastructure-view datasets. 
Results show that integrating data from infrastructure sensors achieves an average of 15\% AP higher than single-vehicle 3D detection, and TCLF can alleviate temporal asynchrony problems.

\section*{Acknowledgements} 
We thank Fan Yang, Ruiwen Zhang, Wenyue Wu, and Xiao Wang from Baidu Inc. for the support in data processing. 
We thank Jilei Mao, Taohua Zhou, Yingjuan Tang, Zan Mao, and Zhiwen Yang for their support in the benchmark construction. 
Thanks to Beijing High-level Autonomous Driving Demonstration Area, Beijing Connected and Autonomous Vehicles Technology Co., Ltd, Baidu Apollo, and Beijing Academy of Artificial Intelligence for their support throughout the dataset construction and release process.

{\small
\bibliographystyle{ieee_fullname}
\bibliography{egbib}
}

\end{document}